\documentclass[10pt,twocolumn,letterpaper]{article}

\usepackage[pagenumbers]{cvpr} 

\usepackage{graphicx}
\usepackage{amsmath}
\usepackage{amssymb}
\usepackage{booktabs}
\usepackage[table]{xcolor}

\usepackage[pagebackref,breaklinks,colorlinks]{hyperref}

\usepackage[capitalize]{cleveref}
\crefname{section}{Sec.}{Secs.}
\Crefname{section}{Section}{Sections}
\Crefname{table}{Table}{Tables}
\crefname{table}{Tab.}{Tabs.}

\def\cvprPaperID{11535} 
\def\confName{CVPR}
\def\confYear{2022}

\begin{document}

\title{{Facial Action Units Detection Aided by Global-Local Expression Embedding}}


\author{
    Zhipeng Hu \textsuperscript{\rm 1}\footnotemark[1],
    Wei Zhang \textsuperscript{\rm 1}\footnotemark[1],
    Lincheng Li \textsuperscript{\rm 1},
    Yu Ding \textsuperscript{\rm 1}\thanks{Equal contribution. Yu Ding is the corresponding author.},
    Wei Chen \textsuperscript{\rm 2},\\
    Zhigang Deng \textsuperscript{\rm 3},
    Xin Yu\textsuperscript{\rm 4} 
     \\ 
 \textsuperscript{\rm 1}Netease Fuxi AI Lab\\
 \textsuperscript{\rm 2}Hebei Agricultural University 
 \textsuperscript{\rm 3}University of Houston \\
 \textsuperscript{\rm 4}University of Technology Sydney \\
 \tt\small \{zphu, zhangwei05, lilincheng, dingyu01\}@corp.netease.com\\
\tt\small rshchchw@hebau.edu.cn; zdeng4@uh.edu; xin.yu@uts.edu.au
}


\maketitle

\begin{abstract}
Since Facial Action Unit (AU) annotations require domain expertise, common AU datasets only contain a limited number of subjects. As a result, a crucial challenge for AU detection is addressing identity overfitting. We find that AUs and facial expressions are highly associated, and existing facial expression datasets often contain a large number of identities. In this paper, we aim to utilize the expression datasets without AU labels to facilitate AU detection. Specifically, we develop a novel AU detection framework aided by the Global-Local facial Expressions Embedding, dubbed GLEE-Net. Our GLEE-Net consists of three branches to extract identity-independent expression features for AU detection. We introduce a global branch for modeling the overall facial expression while eliminating the impacts of identities. We also design a local branch focusing on specific local face regions. The combined output of global and local branches is firstly pre-trained on an expression dataset as an identity-independent expression embedding, and then finetuned on AU datasets. Therefore, we significantly alleviate the issue of limited identities. Furthermore, we introduce a 3D global branch that extracts expression coefficients through 3D face reconstruction to consolidate 2D AU descriptions. Finally, a Transformer-based multi-label classifier is employed to fuse all the representations for AU detection. Extensive experiments demonstrate that our method significantly outperforms the state-of-the-art on the widely-used DISFA, BP4D and BP4D+ datasets.
\end{abstract}

\section{Introduction}
\label{sec:intro}

Facial Action Units (AUs), coded by Facial Action Coding System (FACS), include 32 atomic facial action descriptors based on facial muscle groups~\cite{FACS}. 
Face AU detection has attracted lots of research efforts due to its crucial applications in emotion recognition \cite{AU4EmoctionRecog}, micro-expression detection \cite{AU4MicroExpression}, and mental health diagnosis \cite{AUD4diagnose}.

Since AU annotations require sophisticated expertise and are time-consuming, the size of annotated AU datasets is usually limited, especially in terms of identity variations. \eg, less than 50.
As a result, most AU detection methods overfit to the training identities and do not generalize well to new subjects.
To alleviate the overfitting problem on the small AU datasets, previous methods resort to various auxiliary information as regularization, including facial landmarks~\cite{JPML,li2018eac,shao2021jaa,LP}, unsupervised web images~\cite{zhao2018learning}, emotion priors~\cite{cui2020knowledge}, textual AU descriptions\cite{yang2021exploiting} and so on. 
However, these additional constraints do not directly remove the interference of the training identities from the extracted visual features, thus limiting their performance. 


Different from the previous works, we aim to make the first attempt to employ an expression embedding extracted from the in-the-wild expression dataset ~\cite{vemulapalli2019compact} without AU labels. 
The embedding can provide a strong prior for AU detection due to the two important properties: continuity and identity-independence. 
First, the embedding provides a continuous space for representing the fine-grained expressions. It is beneficial for AU detection since AUs usually show slight variations on the face. 
Second, the embedding is less sensitive to identities because the semantic similar expressions of different identities are analogous in the embedding space. 
This important property can be used to alleviate the overfitting problem in AU detection.
Hence, our motivation is to leverage a continuous expression embedding space to represent AUs for accurate AU detection.

\begin{figure}
    \centering
    \includegraphics[width=1\linewidth]{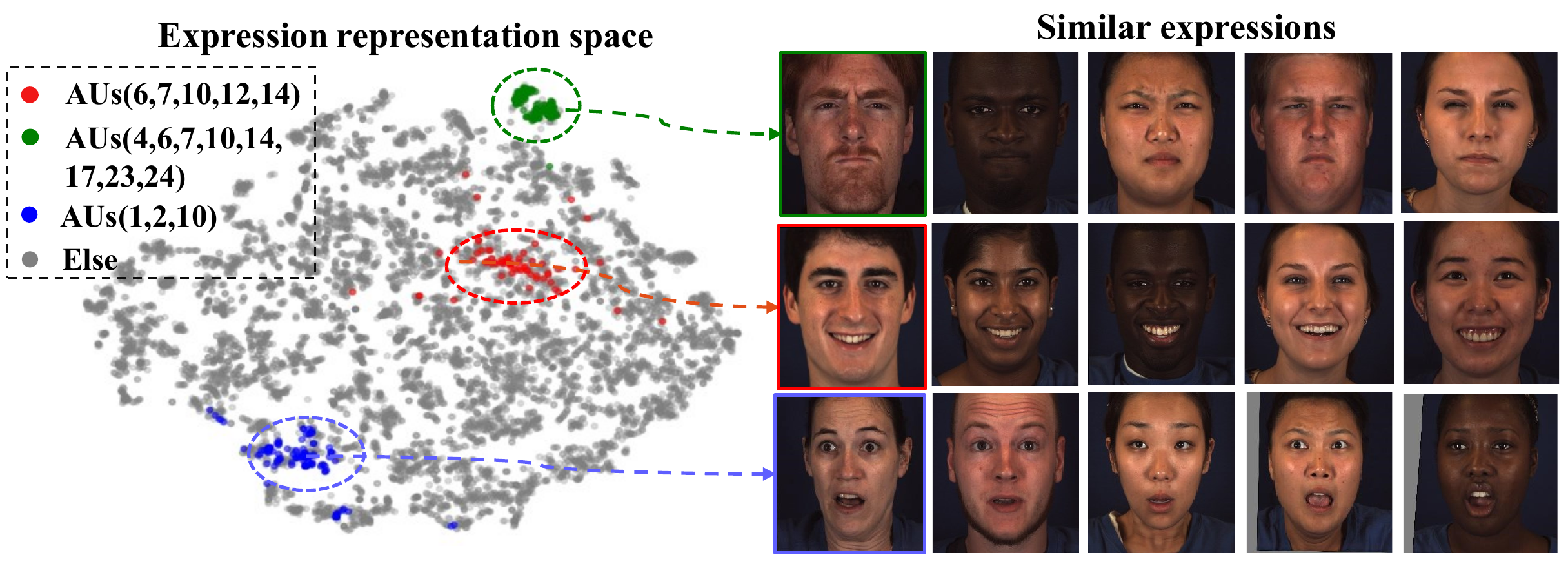}
    \caption{T-SNE visualization of the distributions of some AU combinations in the embedding space of our GLEE-Net. 
    Left: the same AU combinations distribute closely in the expression embedding space.
    Right: similar expressions in the expression embedding space often have similar AU labels and learning expressions can facilitate AU detection.
    }
    \label{fig:query}
\end{figure}

Driven by our motivation, we develop a novel AU detection framework aided by the Global-Local facial Expression Embedding, namely GLEE-Net. Our GLEE-Net consists of three branches that extract identity-independent facial expression features for AU detection. 
To comprehend the overall facial expressions, we introduce a global branch that modeling the expression information as the deviation from the identity representation. 
In this way, our global branch would be less sensitive to the identity information.
Considering AUs are defined on local face regions, we design a local branch to focus on details of specific face regions. 
In order to alleviate the problem of limited identities in AU datasets, different from existing methods with global and local branches, we first pretrain the two branches on an expression dataset \cite{vemulapalli2019compact} and then finetune them on the target AU dataset. 
In this manner, our network has seen various subjects' expressions although their AU labels are not available and moreover, acquires a compact expression embedding for AU detection. 
In Figure~\ref{fig:query}, we sample some images from BP4D and visualize their expression embeddings. As expected, the same AU combinations and similar expressions from different identities are close in our expression embedding space, thus facilitating AU classification. 

Furthermore, in contrast to existing methods that only rely on auxiliary information of 2D images, we find that 3D facial information also provides important expression clues. Thus, we introduce a 3D global branch to obtain expression coefficients, as 3D expression features, through 3D face reconstruction. 
To fully exploit all the representations from our global-local branches, we design a Transformer-based multi-label classifier. Benefiting from the powerful global attention mechanism of Transformer~\cite{vaswani2017attention}, we can effectively fuse different representations and thus explore the correlations among multiple AUs. With the co-occurrence relationships of AUs, our network can predict AUs more accurately.
Extensive experiments demonstrate that our approach significantly achieves superior performance on the widely-used DISFA, BP4D and BP4D+ datasets. 

In summary, the contributions of our work are three-fold:
\begin{itemize}
    \item We propose a novel Global-Local facial Expression Embedding Network (GLEE-Net) for AU detection, which can leverage additional facial expression data (without AU labels) to improve AU detection accuracy.
    
    \item We develop the global and local branches to extract the compact expression embeddings from face regions while paying attention to local facial details.
    To the best of our knowledge, our work is the first attempt to utilize continuous and compact expression features to represent AUs effectively. It achieves appealing generalization capability in addressing AU classification for unseen identities.
    
    \item We introduce a 3D global branch to extract expression coefficients through 3D face reconstruction for AU detection, and demonstrate that exploiting 3D face priors can further improve 2D AU detection.
\end{itemize}

\section{Related Works}

\subsection{AU Detection with Auxiliary Information}
The widely used AU datasets only contain limited subjects due to the difficulty of AU annotation, which is the main cause of overfitting. To resolve it, some works recourse to the various kinds of auxiliary information to enhance the model generalization and facilitate AU detection. 
Introducing extra information of facial landmarks is a common practice in AU detection.
To effectively extract the local features for AUs, JPML~\cite{JPML} utilize the landmarks to crop the facial patches instead of uniformly distributed grids. EAC-Net~\cite{li2018eac} also generate the spatial attention maps according to  the facial landmarks and applies them to the different levels of networks.  
LP-Net~\cite{LP} sents the detected facial landmarks into the P-Net to learn the person-specific shape information.
J{\^A}A-Net~\cite{shao2021jaa} proposes a multi-task framework combining landmark detection and AU detection. 
Besides this, there exist some other kinds of auxiliary information.
Zhao et al.~\cite{zhao2018learning} utilize the unlabelled large-scale web images and propose a weakly-supervised spectral embedding for AU detection.
Cui et al. \cite{cui2020knowledge} construct an expression-AUs knowledge prior 
based on the existing anatomic and psychological research and introduce the expression recognition model for AU detection. 
SEV-Net~\cite{yang2021exploiting} introduces the pre-trained word embedding to learn spatial attention maps based on the textual descriptions of AU occurrences.
The aforementioned methods all directly or indirectly introduce additional data for producing extra regularization in AU detection. We propose to utilize the expression embedding as auxiliary information, which better improves the generalization capability of the AU detection. 

\subsection{AU Detection with global and local features}
Due to the local definition of AUs, many methods attempt to combine the full and regional facial features for AU detection. These works can be classified into three categories: patch-based, multi-task, and text-based methods.

Patch-based methods usually crop the full face into patches according to the local definitions of AUs. DSIN~\cite{LP8} crops 5 patches from a full face based on landmarks and feeds them with the full face into networks for learning the global and local features. ROI-Net~\cite{ROI} also designs a prior landmark cropping rule to crop the inner feature maps. These methods usually suffer from performance degradation in the wild due to erroneous landmark estimation.

To facilitate the model with local features, multi-task methods usually combine AU detection with landmark detection~\cite{benitez2017recognition, shao2021jaa} or landmark-based attention map prediction~\cite{jacob2021facial}. 
In this way, models can extract global features from full faces and also focus on local details from the landmarks for better AU detection. However, these methods ignore that landmarks also contain rich identity information~\cite{jannat2020subject} which may aggravate the identity overfitting.

SEV-Net~\cite{yang2021exploiting} proposes to utilize the textual descriptions of local details to generate a regional attention map. In this way, it highlights the local parts of the global features. However, it requires the extra annotations for the descriptions. 
In addition, the global features of the previous works do not take the removal of the identity disturbance into account. 

Different from the above works, our carefully-designed global branch is dedicated to eliminating identity disturbance, and our cropped patches for the local branch are based on positioned face patches instead of landmarks. 




\subsection{Expression Representations}
The action units reflect the facial expression information and the model perception of the expression plays a crucial role in AU detection. The expression representation can be used to evaluate the expression perception capability of model. 
A common practice to represent expression is mapping the face images into a low-dimensional manifold, which describes the expressions without disturbance of identity, pose or illumination.  
Early works utilize the hidden features of the last or penultimate layer of the model trained in discrete expression classification tasks~\cite{mollahosseini2017affectnet,wang2020region,zhao2021robust} as the expression representation, in which the extracted expression information reflects more information of the limited expression categories but neglect the complicated and fine-grained facial expressions.
Different from them, a compact and continuous embedding for representing facial expressions is proposed by Vemulapalli and Agarwala\cite{vemulapalli2019compact}. It constructs a large-scale facial dataset annotated with expression similarity in a triplet way. Through a large number of triplet comparisons, the trained expression embedding can perceive slight expression changes.
To further reduce the identity influence, Zhang et al.\cite{zhang2021learning} develop a Deviation Learning Network (DLN) with a two-branch structure to achieve more compact and smooth expression embedding.
3D Morphable Model (3DMM) \cite{paysan20093d,booth2018lsfm} has been proposed to fit identities and expression parameters from a single face image. 
Expressions are represented as the coefficients of predefined blendshapes in 3DMM.
The estimated expression coefficients are then used for talking head synthesis \cite{li2021write,zhang2021flow}, expression transfer \cite{yao2021one,kim2018deep} or face manipulation \cite{geng20193d}

\begin{figure*}
    \centering
    \includegraphics[ width=1\linewidth]{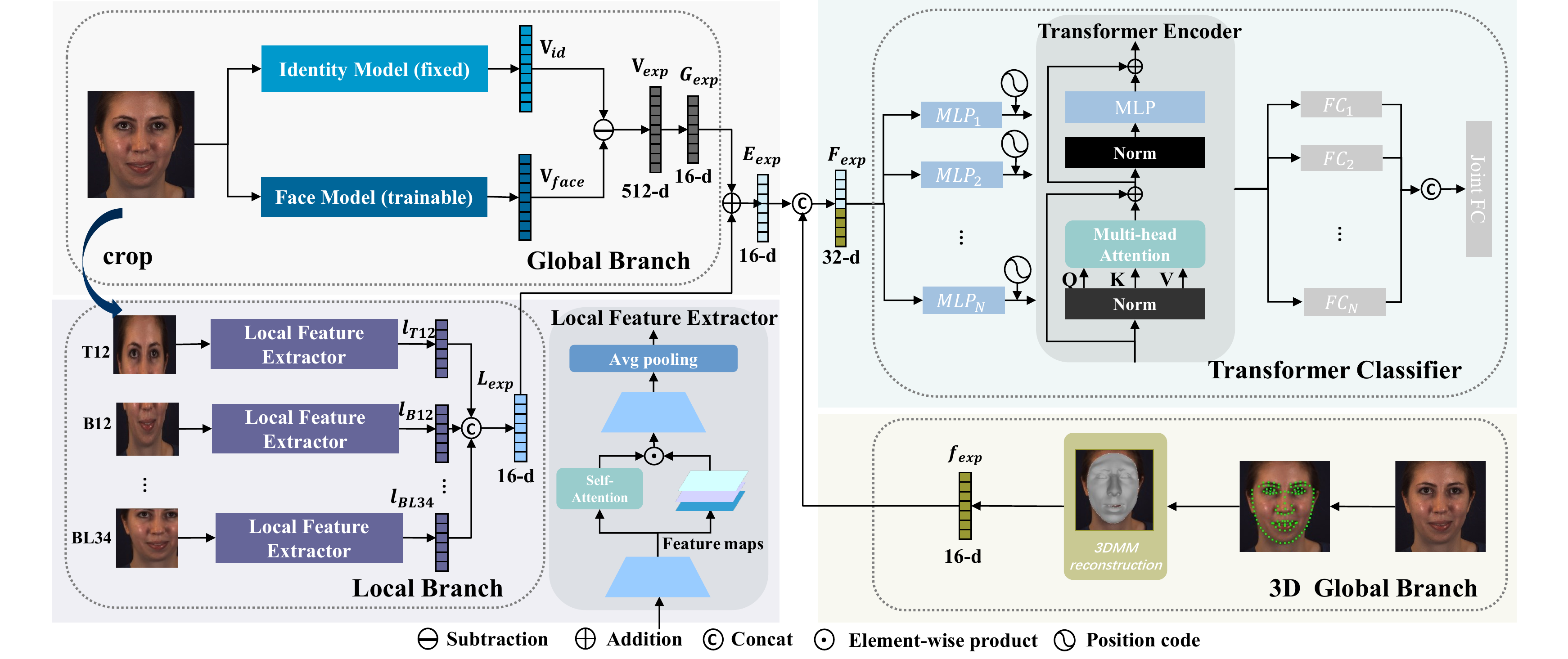}
    \caption{Pipeline of our proposed AU detection framework GLEE-Net. The global and local branches extract the full and partial facial information and are pretrained on the FEC dataset to provide an effective expression prior knowledge for AU detection. To further enhance the expression representation, we exploit the expression coefficients from a 3D global branch. The Transformer classifier models the diverse correlations between AUs and then predicts AUs.}
    \label{fig:pipeline}
\end{figure*}


\section{Proposed Method}

The architecture of the proposed GLEE-Net is shown in Figure~\ref{fig:pipeline}, which takes an image as input and outputs a binary vector to indicate the occurrence of each AU.
The whole framework consists of a global branch, a local branch, a 3D global branch and a Transformer classifier. 
The global branch extracts the full face feature to model the full face expression while the local branch focuses on detailed local information. 
The two branches are pretrained on the FEC expression dataset~\cite{vemulapalli2019compact} and then finetuned on the AU dataset to alleviate the issue of limited identities.
To further enrich 2D facial representations, the 3D global branch extracts the expression coefficients through 3D face reconstruction.
Finally, the Transformer classifier carries out the final AU detection from the combined features of three branches with the powerful attention mechanism.

\subsection{Global Branch}

Inspired by DLN~\cite{zhang2021DLN}, the global branch models the expression feature vector $V_{exp}$ as the deviation from the identity vector $V_{id}$. 
Specifically, the global branch consists of two siamese models, \ie, the face model and the identity model.
The identity model and the face model are initialized with the pretrained FaceNet~\cite{schroff2015facenet} for a face recognition task~\cite{cao2018vggface2}.
Then, we fix the identity model and train the face model to learn the expression deviation. The extracted full face expression feature vector $V_{exp}$ is obtained by:
\begin{equation}
    V_{exp}=V_{face}-V_{id}.
\end{equation}
The deviation model of the global branch benefits from an effective feature initialization that can alleviate the disturbance of expression-irrelevant information, such as identity, pose, etc. After a linear layer for dimension reduction, we obtain $G_{exp}$ as the global expression feature vector.


\subsection{Local Branch}

\begin{figure}
    \centering
    \includegraphics[ width=0.8\linewidth]{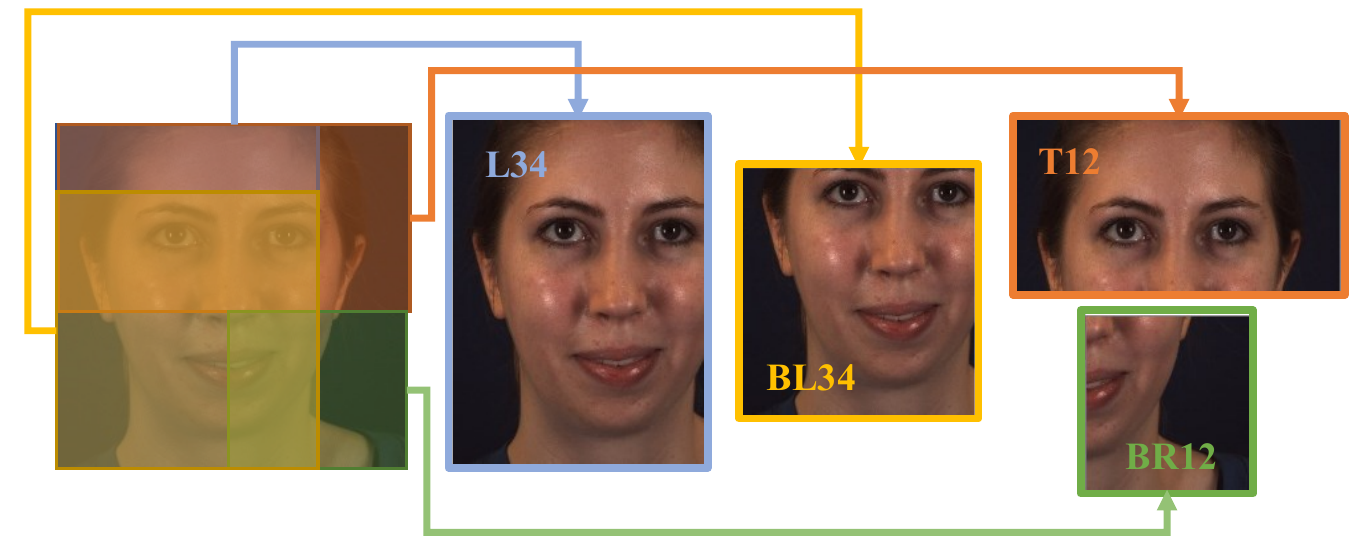}
    \caption{Illustration of cropped facial parts. Four of the 16 specific parts used in our method are shown (L34, BL34, T12 and BR12).}
    \label{fig:crop_rule}
\end{figure}

{We introduce a local branch to facilitate the global branch with more detailed local information, which is also beneficial for AU detection due to the local nature of AUs.}
First, we crop the image into 16 parts for local part extraction. Since the expression dataset contains a large number of in-the-wild images, it is hard to locate specific face regions accurately. 
Therefore, we choose to crop the image according to the whole image area instead of facial landmarks. 
Specifically, we crop three-quarters of the image from left, right, top and bottom and call them L34, R34, T34 and B34 respectively. Similarly, we crop half from the left, right, top and bottom as L12, R12, T12 and B12 respectively. We also crop the image from two directions simultaneously to construct eight local parts as TL34, TR34, TL12, TR12, BL34, BR34, BL12 and BR12 respectively. Figure~\ref{fig:crop_rule} shows some examples of our cropping strategy. We can observe that different parts focus on different face regions~(\eg, T12 focuses on eyebrows and eyes).


After obtaining the local parts of a face, we resize them to $96\times96$ and feed them into a group of local feature extractors separately. The structure of the local feature extractor is illustrated in Figure~\ref{fig:pipeline}. Specifically, after four $3\times3$ convolutional layers, a self-attention module is employed to strengthen the importance of crucial local positions on a face. Afterwards, the feature maps are resized through average pooling. We concatenate the local features of the 16 face regions and resize the final local expression feature $L_{exp}$ to 16 by MLP layers. We obtain the final expression embedding $E_{exp}$ by adding $L_{exp}$ and the global expression feature $G_{exp}$. $E_{exp}$ has the comprehensive expression representation capability from the global-local 2D face regions. 


\subsection{3D Global Branch}
To further enrich the expression embedding, the 3D global branch estimates the expression coefficients from a 3DMM. {We find 3D expression coefficients are generally identity-independent and our work is the first to introduce it as another input feature for AU detection. }

Given the face shape coefficients \(f_s\in \mathbb{R}^{N_{s}}\) and the expression coefficients \(f_{exp}\in \mathbb{R}^{N_{e}}\), the parametric 3D mesh \(M(s,e)\) is represented as:
\begin{equation}
M(s, e) = M_{0} +\sum_{i=1}^{N_{s}}S_{i}\cdot (f_{s})_{i} + \sum_{j=1}^{N_e}E_{j}\cdot (f_{exp})_{j},
\end{equation}
where \(M_{0}\) is the mean face mesh,  and \(\{S_i\}_{i=1}^{N_s}\) and
\(\{E_i\}_{i=1}^{N_e}\) are the linear shape and expression bases, respectively. 
We take \(M_{0}\) and \(\{S_i\}_{i=1}^{N_s}\) with $N_s=60$ from LSFM \cite{booth2018lsfm} and normalize them into the range of \([-3,3]\). 
We sculpt $N_e=51$ blendshapes on LSFM as \(\{E_i\}_{i=1}^{N_e}\).

To reconstruct \(f_{exp}\) from the input image, we perform 3DMM fitting by minimizing the energy function:
\begin{equation}
\begin{aligned}
\label{eqn:3DMM_Opt}
\mathcal{L}_{3D} = \sum_{i=1}^{N_l}&D(p_i, P(M(f_{s},f_{exp})_i,\xi) \\ &+ \lambda_e \left \| f_{exp} \right \|_{2}^{2} + \lambda_s \left \| f_{s} \right \|_{2}^{2},
\end{aligned}
\end{equation}
where \(N_l=68\) is the number of facial landmarks, \(p_i\) is the pixel coordinate of the detected \(i\)-th landmark, and \(M(s,e)_i\) is the 3D vertex of the \(i\)-th landmark on mesh \(M\). $P$ is the perspective projection matrix \cite{andrew2001multiple} that projects a 3D vertex into its homogeneous pixel coordinate according to the head pose $\xi$.
The first term \(D(\cdot, \cdot)\) measures the distance between a detected image landmark and its corresponding projected mesh landmark.
Since the contour landmark (\eg, lip contours) distributions of the image detector and 3D mesh notation are not perfectly aligned, for non-endpoint landmarks, we define the point-to-curve pixel distance between the projected 3D landmark point and the detected 2D face contour curve. 
The second and third regularization terms with weights \(\lambda_e=10^{-4}\) and \(\lambda_s=10^{-4}\) respectively.
We employ the Levenberg-Marquard (LM) algorithm \cite{kelley1999iterative} to solve the best $f_{exp}$ and $f_s$.

After optimization, the estimated expression coefficients $f_{exp}$ are concatenated with the global-local expression embedding $E_{exp}$ for AU classification. $f_{exp}$ is not applied in the pretraining process of expression embedding  because the quality of in-the-wild FEC~\cite{vemulapalli2019compact} images is often worse than that of AU detection images and may lead to estimation errors in 3D reconstruction.

\subsection{Transformer Classifier}
We notice that the correlation among multiple AUs provides additional constraints and thus facilitates AU predictions.
Therefore, rather than a conventional multi-class classifier, we model the correlation among AUs by Transformer encoders~\cite{vaswani2017attention}. 
In this way, the prediction of the occurrence of each AU takes into account the correlations of all the other AUs.
{As shown in Figure~\ref{fig:pipeline}, the parallel MLP layers first extract multi-view information related to AUs from the joint expression representation $F_{exp}$.  
Then, three Transformer encoders take the extracted multi-view AU features as an input sequence. 
Based on the multi-head attention mechanism, the Transformer encoders can fully exploit the latent correlations among AUs.}
Finally, the enhanced features are passed through independent FC layers as initial predictions and a joint FC layer is employed to fuse the initial predictions to achieve a more accurate final prediction.

\subsection{Network Training}
\textbf{Pretraining for expression embedding:}
The global and local branches are pretrained on the expression dataset FEC~\cite{vemulapalli2019compact} with a triplet loss to obtain a compact expression embedding $E_{exp}$ as an auxiliary task. 
FEC is a large-scale in-the-wild expression dataset that contains 500,203 triplets annotated by human perception of expression. 
One triplet consists of one Anchor~(A), Positive~(P) and Negative~(N). A and P have more similar expressions than N.
Following such annotation, we use the triplet loss to pretrain the two branches as follows:
\begin{equation}
\begin{aligned}
\mathcal{L}_{tri}  &=\max(0,\|E^{A}_{exp}-E^{P}_{exp}\|_2^2-\|E^{A}_{exp}-E^{N}_{exp}\|_2^2+m) \\
          &+\max(0,\|E^{A}_{exp}-E^{P}_{exp}\|_2^2-\|E^{P}_{exp}-E^{N}_{exp}\|_2^2+m),
\end{aligned}
\end{equation}
where $E^{(.)}_{exp}$ represents the corresponding expression embedding after normalization, and $m$ is the margin.  
Due to a large number of expression comparisons from different identities in FEC, the extracted expression embeddings can be identity-independent and continuous. Also, the expression representation can be more robust with the local branch since the cropped parts provide rich detailed local information. The expression embedding is of great value for the generalization of AU detection.

\noindent\textbf{Training for AU detection:}
Aided by the effective expression embeddings, we finetune the whole network for the AU detection task. Specifically, we employ the weighted cross entropy loss to train our AU classifiers, defined by:
\begin{equation}
\begin{aligned}
\mathcal{L}_{AU}=-\frac{1}{N_{a}}\sum_{i=1}^{N_{a}}{\frac{1}{r^{i}}[g^i\log{p^i}+(1-g^i)\log(1-p^i)]},
\end{aligned}
\label{eq:lossfunc}
\end{equation}
where $N_a$ is the number of AUs, $g^i$ is the ground-truth binary label of the $i$-th AU, $p^i$ is the predicted value, and $r^i$ is the prior occurrence ratio of the $i$-th AU in the training set to balance the weight of different AUs.
The loss is applied on both the initial predictions before joint FC and the final predictions.
In the inference stage, the $i$-th AU is determined as presence if $p^i>0.5$.

\begin{table*} 
\footnotesize
\renewcommand\arraystretch{1.05} 
\centering
{
  \caption{Comparisons of our method and the state-of-the-art methods on BP4D in terms of F1 scores (\%). The best results are shown in bold, and the second best are in brackets.}
  \vspace{-1em}
  \label{tab:F1 on Bp4D}
  \begin{tabular}{l|cccccccccccc|c}
    \hline 
   \rowcolor{gray!25} \textbf{Method} &\textbf{AU1} &\textbf{AU2} &\textbf{AU4} &\textbf{AU6} &\textbf{AU7} &\textbf{AU10} &\textbf{AU12} &\textbf{AU14} &\textbf{AU15} &\textbf{AU17} &\textbf{AU23} &\textbf{AU24} &\textbf{Avg.}\\[1pt]
    \hline \hline 
    LSVM    &23.2 &22.8 &23.1 &27.2 &47.1 &77.2 &63.7 &64.3 &18.4 &33.0 &19.4 &20.7 &35.3 \\
    JPML    &32.6 &25.6 &37.4 &42.3 &50.5 &72.2 &74.1 & [65.7] &38.1 &40.0 &30.4 &42.3 &45.9\\
    DRML    &36.4 &41.8 &43.0 &55.0 &67.0 &66.3 &65.8 &54.1 &33.2 &48.0 &31.7 &30.0 &48.3\\
    EAC-Net &39.0 &35.2 &48.6 &76.1 &72.9 &81.9 &86.2 &58.8 &37.5 &59.1 &35.9 &35.8 &55.9\\
    DSIN    &51.7 &40.4 &56.0 &76.1 &73.5 &79.9 &85.4 &62.7 &37.3 &62.9 &38.8 &41.6 &58.9\\
    ARL     &45.8 &39.8 &55.1 &75.7 &77.2 &82.3 &86.6 &58.8 &47.6 &62.1 &47.4 &55.4 &61.1\\
    SRERL   &46.9 &45.3 &55.6 &77.1 &78.4 &83.5 &87.6 &63.9 &52.2 &[63.9] &47.1 &53.3 &62.9\\
    LP-Net  &43.4 &38.0 &54.2 &77.1 &76.7 &83.8 &87.2 &63.3 &45.3 &60.5 &48.1 &54.2 &61.0\\
    UGN-B   &54.2 &46.4 &56.8 &76.2 &76.7 &82.4 &86.1 &64.7 &51.2 &63.1 &[48.5] &53.6 &63.3\\
    J{\^A}A-Net &53.8 &47.8 &58.2 &78.5 &75.8 &82.7 &\textbf{88.2} &63.7 &43.3 &61.8 &45.6 &49.9 &62.4\\
    MHSA-FFN &51.7 &[49.3] &\textbf{61.0} &77.8 &\textbf{79.5} &82.9 &86.3 & \textbf{67.6} &51.9 &63.0 &43.7 &[56.3] &[64.2]\\
    SEV-Net &[58.2] &\textbf{50.4} &58.3 &\textbf{81.9} &73.9 &\textbf{87.8} &87.5 &61.6 &[52.6] &62.2 &44.6 &47.6 &63.9\\
    HMP-PS &53.1 &46.1 &56.0 &76.5 &76.9 &82.1 &86.4 &64.8 &51.5 &63.0 &\textbf{49.9} &54.5 &63.4\\
    \hline
    GLEE-Net (Ours)  &\textbf{60.6} &44.4 &\textbf{61.0} &[80.6] &[78.7] &[85.4] &[88.1] &64.9 &\textbf{53.7} &\textbf{65.1} &47.7 &\textbf{58.5} &\textbf{65.7}\\
    \hline
  \end{tabular}}
\end{table*}

\begin{table*}[t]
\footnotesize
\renewcommand\arraystretch{1.05} 
\centering
{ 
  \caption{Comparisons of our method and the state-of-the-art methods on DISFA in terms of F1 scores (\%). 
  }
  \vspace{-1em}
  \label{tab:F1 on DISFA}
  \begin{tabular}{l|cccccccc|c}
   \hline 
 \rowcolor{gray!25} \textbf{Method} &\textbf{AU1} &\textbf{AU2} &\textbf{AU4} &\textbf{AU6} &\textbf{AU9} &\textbf{AU12} &\textbf{AU25} &\textbf{AU26} &\textbf{Avg.}\\[1pt]
    \hline\hline
    LSVM       &10.8 &10.0 &21.8 &15.7 &11.5 &70.4 &12.0 &22.1 &21.8\\
    DRML       &17.3 &17.7 &37.4 &29.0 &10.7 &37.7 &38.5 &20.1 &26.7\\
    APL        &11.4 &12.0 &30.1 &12.4 &10.1 &65.9 &21.4 &26.9 &23.8\\
    EAC-Net        &41.5 &26.4 &66.4 &50.7 &8.5 &\textbf{89.3} &88.9 &15.6 &48.5\\
    DSIN       &42.4 &39.0 &68.4 &28.6 &46.8 &70.8 &90.4 &42.2 &53.6\\
    ARL        &43.9 &42.1 &63.6 &41.8 &40.0 &76.2 &[95.2] &66.8 &58.7\\
    SRERL      &45.7 &47.8 &56.9 &47.1 &45.6 &73.5 &84.3 &43.6 &55.9\\
    LP-Net     &29.9 &24.7 &72.7 &46.8 &49.6 &72.9 &93.8 &56.0 &56.9\\
    UGN-B      &43.3 &48.1 &63.4 &49.5 &48.2 &72.9 &90.8 &59.0 &60.0\\
    J{\^A}A-Net    &\textbf{62.4} &\textbf{60.7} &67.1 &41.1 &45.1 &73.5 &90.9 &[67.4] &[63.5]\\
    MHSA-FFN   &46.1 &48.6 &[72.8] &\textbf{56.7} &50.0 &72.1 &90.8 &55.4 &61.5\\
    SEV-Net    &55.3 &53.1 &61.5 &[53.6] &38.2 &71.6 &\textbf{95.7} &41.5 &58.8\\
    HMP-PS     &38.0 &45.9 &65.2 &50.9 &[50.8] &76.0 &93.3 &\textbf{67.6} &61.0\\
    \hline
    GLEE-Net (Ours)    &[61.9] &[54.0] &\textbf{75.8} &45.9 &\textbf{55.7} &[77.6] &92.9 &60.0 &\textbf{65.5}\\
    \hline
  \end{tabular}}
\end{table*}

\section{Experiments}
\subsection{Datasets}
We evaluate our method on three widely-used AU detection datasets: BP4D \cite{BP4D}, DISFA \cite{MavadatiDISFA}, and BP4D+ \cite{zhang2016multimodal}.

\textbf{BP4D} is composed of about 140,000 frames from 328 video clips. Each video shows the spontaneous reactions of a subject when he is doing a specific task. The subjects contain 23 females and 18 males. The annotation of each frame includes the occurrence or absence of 12 AUs and the intensity of 5 AUs. For fair comparisons, we choose the same three-fold dataset division as the previous works\cite{li2018eac, shao2021jaa} and perform the three-fold cross validation.

\textbf{DISFA} is composed of about 130,000 frames from 27 video clips. Like BP4D, the expressions in videos come from human spontaneous reactions. The subjects contain 12 females and 15 males, which is less than BP4D. The annotation of each frame includes the intensities~(range from 0 to 5) of 8 AUs. Following previous works, the intensity with $\{0,1\}$ value represents the AU absence and the intensity with $\{2,3,4,5\}$ value represents the AU occurrence. We also compare our method with other works based on the three-fold cross validation.

\textbf{BP4D+} extends more subjects based on BP4D. It contains 1400 video clips from 140 subjects (82 females and 58 males). About 198,000 frames are annotated with the occurrence or absence of the same 12 AUs in BP4D. Following the previous works, we also perform three-fold cross validation. 
Moreover, to further prove the superior generalization of our method, we also construct the experiment that is trained on BP4D and evaluated on BP4D+ following the works \cite{attnetionTrans2019,shao2021jaa}.

\subsection{Implementation Details}


We first pretrain the global and local branches on FEC~\cite{vemulapalli2019compact} and then finetune the whole framework on evaluated AU datasets. 
In the pretraining stage, we use an SGD optimizer with momentum 0.9 and learning rate $2\times10^{-4}$. The pretraining converges within 10 epochs with a batch size of 30. 
In the finetuning stage, we train the entire framework with an SGD optimizer with momentum 0.9 and a learning rate $2\times10^{-3}$ for 10 epochs. We employ the method \cite{baltrusaitis2018openface} as the facial landmark detector. The input face images are aligned by similarity transformation as J{\^A}A-Net \cite{shao2021jaa}. 
The computational cost of our method involves 8.72 GFLOPs. The number of parameters is 56.6M. Our GLEE-Net is not heavy-weighted because our local feature extractor for per patch only contains a few layers, and the input of the Transformer encoder is 12 AU features of dimension 32.

\subsection{Comparison with the State-of-the-Art}
We first compare our method with the state-of-the-art. The compared methods include LSVM \cite{LSVM}, JPML \cite{JPML}, DRML \cite{DRML}, APL \cite{LP48}, EAC-Net \cite{li2018eac}, DSIN \cite{LP8}, ARL \cite{attnetionTrans2019}, SRERL \cite{SRERL}, ML-GCN \cite{chen2019multi}, MS-CAM \cite{you2020cross}, LP-Net \cite{LP}, FACS3D-Net \cite{yang2019facs3d}, UGN-B \cite{song2021uncertain}, J{\^A}A-Net \cite{shao2021jaa}, MHSA-FFN \cite{jacob2021facial}, SEV-Net \cite{yang2021exploiting}, HMP-PS \cite{song2021hybrid}. 
Following the experimental settings of most previous works, 
we employ the F1 score \cite{F1} on the frame level as the evaluation metric. 

\textbf{Evaluation on BP4D.}
The results on BP4D are shown in Table \ref{tab:F1 on Bp4D}. The average F1 score of our method is 65.7\%, which outperforms all the competing methods. In terms of the performance on single AUs, our GLEE-Net achieves the highest or the second highest F1 score among all the methods on 9 of the 12 evaluated AUs. Compared to SEV-Net and J{\^A}A-Net that employ textual embedding and facial landmarks as auxiliary training information, our method obtains better results with the compact expression embedding as prior knowledge. Our method also outperforms MHSA-FFN that also introduces Transformer into classifiers.

\textbf{Evaluation on DISFA.}
Table \ref{tab:F1 on DISFA} shows the results on DISFA. The average F1 score of our method outperforms all the state-of-the-art methods. Compared to the most recent works SEV-Net, HMP-PS and MHSA-FFN, our method obtains the improvement of 11.4\%, 7.4\% and 6.5\%, respectively. As the number of subjects in DISFA is smaller than that of BP4D, most of the previous methods suffer from overfitting to the appearance of training subjects more severely. Therefore, benefiting from our identity-independent expression representations, our method achieves an even more significant improvement. 

\begin{table*}[!t]
\footnotesize
\renewcommand\arraystretch{1.05} 
\centering
{
  \caption{Comparisons of our method and the state-of-the-art methods on BP4D+ in terms of F1 scores (\%). 
  }
 \vspace{-1em}
  \label{tab:F1 on ThreeFoldBp4DPlus}
  \begin{tabular}{l|cccccccccccc|c}
    \hline
   \rowcolor{gray!25} \textbf{Method} &\textbf{AU1} &\textbf{AU2} &\textbf{AU4} &\textbf{AU6} &\textbf{AU7} &\textbf{AU10} &\textbf{AU12} &\textbf{AU14} &\textbf{AU15} &\textbf{AU17} &\textbf{AU23} &\textbf{AU24} &\textbf{Avg.}\\[1pt]
    \hline \hline
    FACS3D-Net     &43.0 &38.1 &\textbf{49.9} &82.3 &85.1 &87.2 &87.5 &66.0 &48.4 &[47.4] &50.0 &[31.9] &59.7\\
    ML-GCN &40.2 &36.9 &32.5 &84.8 &[88.9] &89.6 &[89.3] &81.2 &[53.3] &43.1 &55.9 &28.3 &60.3\\
    MS-CAM &38.3 &37.6 &25.0 &85.0 &\textbf{90.9} &\textbf{90.9} &89.0 &[81.5] &\textbf{60.9} &40.6 &[58.2] &28.0 &60.5\\
    SEV-Net &[47.9] &[40.8] &31.2 &\textbf{86.9} &87.5 &89.7 &88.9 &\textbf{82.6} &39.9 & \textbf{55.6} &\textbf{59.4} &27.1 &[61.5]\\
    \hline
    GLEE-Net (Ours)  &\textbf{54.2} &\textbf{46.3} &[38.1] &[86.2] &87.6 &[90.4] &\textbf{89.5} &81.3 &46.3 & [47.4] &57.6 &\textbf{39.6} &\textbf{63.7}\\
    \hline
  \end{tabular}}
\end{table*}

\begin{table*}[!t]
\footnotesize
\renewcommand\arraystretch{1.05} 
\centering
{
  \caption{Comparisons of our method and the state-of-the-art on cross-dataset evaluation (trained on BP4D and evaluated on BP4D+) in terms of F1 scores (\%).}
  \vspace{-1em}
  \label{tab:F1 on Bp4DPlus}
  \begin{tabular}{l|cccccccccccc|c}
    \hline
    \rowcolor{gray!25}\textbf{Method} &\textbf{AU1} &\textbf{AU2} &\textbf{AU4} &\textbf{AU6} &\textbf{AU7} &\textbf{AU10} &\textbf{AU12} &\textbf{AU14} &\textbf{AU15} &\textbf{AU17} &\textbf{AU23} &\textbf{AU24} &\textbf{Avg.}\\[1pt]
    \hline \hline
    EAC-Net &38.0 &[37.5] &[32.6] &82.0 &83.4 &87.1 &85.1 &62.1 &44.5 &43.6 &45.0 &[32.8] &56.1\\
    ARL     &29.9 &33.1 &27.1 &81.5 &83.0 &84.8 &86.2 &59.7 &[44.6] &[43.7] &[48.8] &32.3 &54.6\\
    J{\^A}A-Net &[39.7] &35.6 &30.7 &[82.4] &[84.7] &[88.8] &[87.0] &[62.2] &38.9 &\textbf{46.4} &\textbf{48.9} &\textbf{36.0} &[56.8]\\
    \hline
    GLEE-Net (Ours)  &\textbf{39.8} &\textbf{37.9} &\textbf{41.6} &\textbf{83.4} &\textbf{88.2} &\textbf{90.2} &\textbf{87.4} &\textbf{76.6} &\textbf{48.3} &42.9 &47.7 &29.8 &\textbf{59.5}\\
    \hline
  \end{tabular}}
\end{table*}

\begin{table*}[!t]
\footnotesize
\renewcommand\arraystretch{1.05} 
\centering
{
  \caption{Ablation study on BP4D measured by F1 scores (\%). }
  \vspace{-1em}
  \label{tab:F1 on Ablation}
 \begin{tabular}{l|cccccccccccc|c}
    \hline
   \rowcolor{gray!25}\textbf{Method} &\textbf{AU1} &\textbf{AU2} &\textbf{AU4} &\textbf{AU6} &\textbf{AU7} &\textbf{AU10} &\textbf{AU12} &\textbf{AU14} &\textbf{AU15} &\textbf{AU17} &\textbf{AU23} &\textbf{AU24} &\textbf{Avg.}\\[1pt]
    \hline\hline
    
    w/o pretrain &54.8 &44.5 &52.3 &77.6 &75.9 &82.5 &86.7 &63.6 &45.6 &62.3 &47.5 &51.0 &62.0\\
    w. fixed GB \& LB &55.1 &44.9 &49.1 &76.5 &75.9 &82.2 &87.2 &65.0 &36.3 &59.8 &34.4 &50.2 &59.7\\
    \hline
    w. ResNet &55.0 &45.2 &52.0 &77.8 &77.8 &83.3 &[87.8] & \textbf{65.8} &46.2 &63.8 &45.8 &53.5 &62.8\\
    w/o GB \& LB  &55.2 &45.2 &46.1 &76.1 &75.6 &81.5 &86.0 &64.8 &36.0 &59.7 &34.1 &49.4 &59.1\\
    w/o 3DGB &54.4 &43.3 &59.2 &[79.5] & [78.1] & [84.7] &87.6 &64.6 &[53.9] &63.9 &47.3 &[58.2] &64.6\\
    w/o LB &58.9 &[46.8] &55.9 &78.3 &78.0 &83.3 &87.6 &62.1 &52.8 &[64.2] &\textbf{50.3} &53.5 &64.3\\
    \hline
    w/o TC & [59.7] &43.0 &\textbf{64.0} &78.7 &77.2 &83.3 &87.5 &63.8 &\textbf{54.3} &[64.2] &46.7 &56.0 & [64.9]\\
    w/o joint FC &56.3 & \textbf{47.3} &57.0 &77.9 &76.1 &84.0 &87.7 &[65.4] &51.6 &63.2 & [48.9] &56.5 &64.3\\
    \hline
    GLEE-Net (Full) & \textbf{60.6} &44.4 & [61.0] & \textbf{80.6} & \textbf{78.7} &\textbf{85.4} & \textbf{88.1} &64.9 &53.7 & \textbf{65.1} &47.7 & \textbf{58.5} &\textbf{65.7}\\
    \hline
  \end{tabular}}
\end{table*}

\textbf{Evaluation on BP4D+.}
Table \ref{tab:F1 on ThreeFoldBp4DPlus} shows the results of the three-fold cross validation on BP4D+. Results of the compared methods are reported by SEV-Net \cite{yang2021exploiting}. Our method achieves the F1 score of 63.7\% and outperforms all the state-of-the-art methods.
To further evaluate the generalization performance on large scale testing identities, we train our method on BP4D and evaluate it on the full BP4D+ of 140 subjects. The results are shown in Table \ref{tab:F1 on Bp4DPlus}. We use the reported results in the work \cite{shao2021jaa} for comparison. Again, our method outperforms all the compared methods under the large-scale cross-dataset evaluation. It proves that our method can extract identity-independent information and generalize well to new identities.

\begin{figure*}[!t]
    \centering
    \includegraphics[ width=1\linewidth]{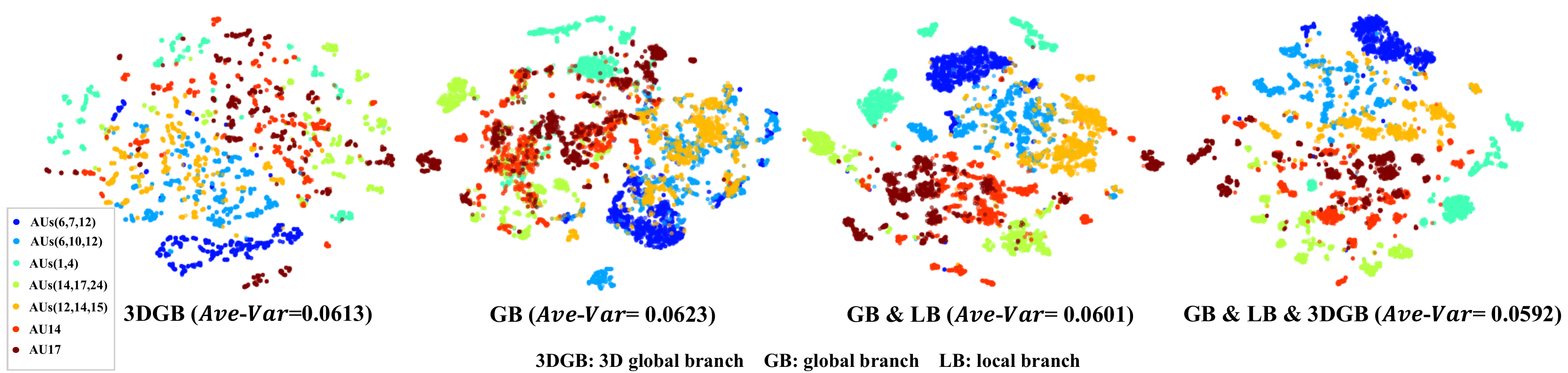}
    \caption{T-SNE visualization of sampled AU combinations in the four different representation spaces. 
    It is observed that representations from our proposed GLEE-Net~(GB \& LB \& 3DGB) are more compact in each AU combination. \textit{Ave-Var} is the average of representation variances of all the AU combinations.
    }
    \label{fig:tsne}
\end{figure*}
\begin{figure*}[!t]
    \centering
    \includegraphics[ width=1\linewidth]{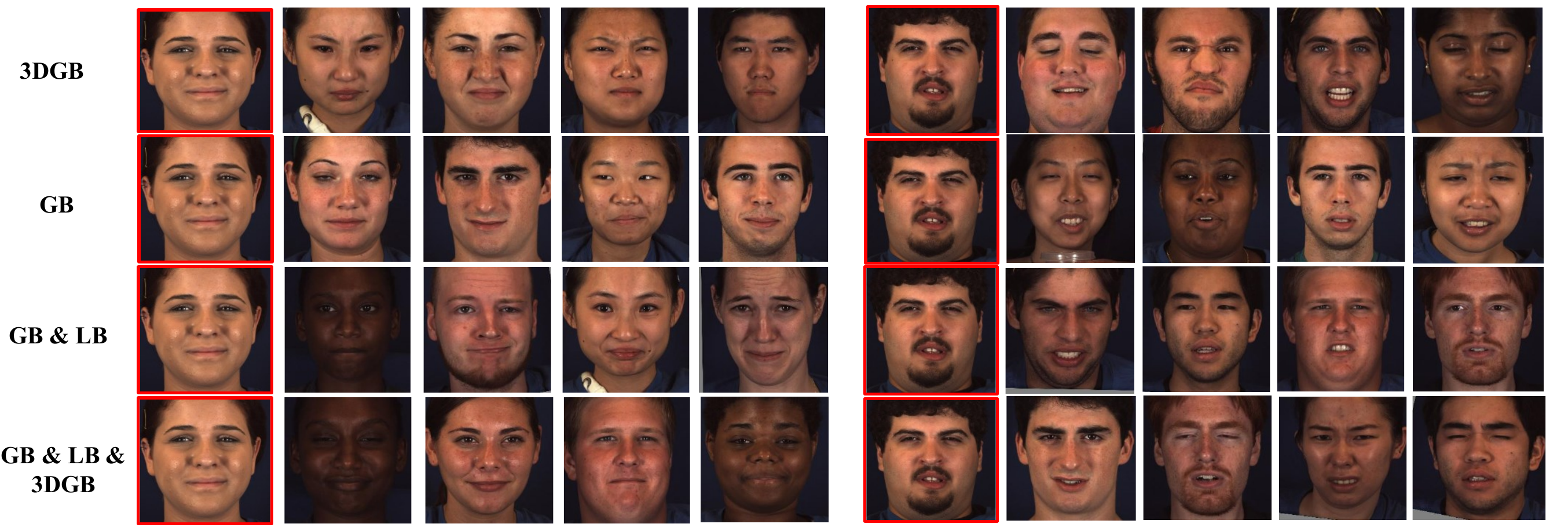}
    \caption{Illustration of our expression representation facilitating AU detection. We attain one image as a center (marked in red) and sample its surrounding features from the expression representation space. It can be seen that images with similar AUs are retrieved in this space.}
    \label{fig:query2}
\end{figure*}


\subsection{Ablation Study}
We also conduct an ablation study to evaluate the effectiveness of each module by removing or replacing it with a baseline method. The results are shown in Table \ref{tab:F1 on Ablation}.

\textbf{Prior Knowledge. }
In the experiment without pretraining on the expression dataset~(w/o pretrain), we can see an apparent drop in average F1. It proves the significance of prior knowledge from the compact expression embedding. We also set up an experiment with fixed pretrained parameters of global and local branches~(w. fixed GB \& LB) and obtain a lower average F1, which indicates a domain gap exists between FEC and BP4D datasets.

\textbf{Expression representation. }
To validate the effectiveness of representations extracted by the 3D global, global and local branches, we construct the model with only 3D global branch~(w/o GB \& LB) and only the global and local branches~(w/o 3DGB) individually. 
To indicate the significance of the global and local branches, we replace these two branches with a ResNet~\cite{Resnet34}~(w. ResNet).
Also, we build up a model without the local branch~(w/o LB) to confirm the usefulness of the local branch.
The experiments show removing those components lead to degraded performance compared to full GLEE-Net, proving that three branches are all effective. 
The local branch is used to supply extra local detail information to the global branch rather than working individually. 

\textbf{Classifier. }
We replace the Transformer classifier with an MLP classifier (w/o TC), the decrease of average F1 indicates that the Transformer classifier improves AU detection performance. We also remove the joint FC layer of the Transformer classifier and the final prediction (w/o joint FC). The performance drop validates the practical design of our two-stage predictions.


\subsection{Visual Analysis}


To demonstrate the generalization of GLEE-Net via expression representations, we sample some images with seven common AU combinations in BP4D, and show the t-SNE visualization of their expression representations in Figure~\ref{fig:tsne}. We compute the expression representation variance for each AU combination and average them as \textit{Ave-Var}. Samples of the same AU combination from different identities are distributed more closely under our full method~(GB \& LB \& 3DGB) 
than other ablation settings. This indicates that our expression representation focuses on the AU features rather than overfitting to training identities. 


Figure~\ref{fig:query2} shows the expression representation facilitates AU detection. We choose one image as center~(in red) and sample its surrounding features in the representation space. 
Under the joint features of the three branches~(GB \& LB \& 3DGB), neighbouring images all have the same facial expressions. The other three versions roughly capture similar facial actions, but there are still subtle variances of AUs.

\subsection{Discussion}
Existing top-ranked works like MHSA-FFN~\cite{jacob2021facial} and SEV-Net~\cite{yang2021exploiting} also use additional dataset information, such as landmarks, word embedding, and so on, but they still suffer from the identity overfitting problem on DISFA that contains smaller subjects. 
Compared with previous works, our method is the first to introduce the effective expression embedding and 3D expression coefficients to alleviate the issue of overfitting caused by limited identities of AU datasets. 
Our method makes full of the identity-independent expression representation based on additional dataset information~(without AU labels) to facilitate AU detection. 

Compared with MHSA-FFN and SEV-Net, our method achieves the promising results on widely-used BP4D, BP4D+ and DISFA datasets. Also, cross-dataset evaluation demonstrates our better generalization than the state-of-the-art methods. 

In our future works, we intend to investigate in-the-wild AU datasets such as SEWA~\cite{kossaifi2019sewa} and Aff-wild2~\cite{kollias2018aff}, which have not been widely employed in existing works.


\section{Limitations}
While our method outperforms the state-of-the-art methods on the 12 or 8 evaluated AUs, it cannot detect the rest of AUs due to the lack of annotated data. Besides, our method may suffer an accuracy decrease under the complex lighting and occlusion environment due to the limitation of expression embedding and 3D expression coefficients.

\section{Conclusion}

This paper presents a novel AU detection framework, dubbed GLEE-Net, by fully taking advantage of available facial expression data.
Through the global and local branches, our GLEE-Net extracts rich representations of input faces in a compact and continuous expression space. Taking advantage of the expression representations, similar AU combinations cluster closer in the latent space. This not only significantly facilitates AU classification but also makes AU detection generalized well to various subjects. Moreover, our introduced 3D facial expression branch provides complementary information to 2D AU detection. Our Transformer-based classifier effectively fuses representations of three branches and produces more accurate AU detection results than the state-of-the-art. 


{\small
\bibliographystyle{ieee_fullname}
\bibliography{egbib}

\begin{thebibliography}{10}\itemsep=-1pt

\bibitem{andrew2001multiple}
Alex~M Andrew.
\newblock Multiple view geometry in computer vision.
\newblock {\em Kybernetes}, 2001.

\bibitem{baltrusaitis2018openface}
Tadas Baltrusaitis, Amir Zadeh, Yao~Chong Lim, and Louis-Philippe Morency.
\newblock Openface 2.0: Facial behavior analysis toolkit.
\newblock In {\em FG 2018}, pages 59--66. IEEE, 2018.

\bibitem{benitez2017recognition}
Carlos~Fabian Benitez-Quiroz, Yan Wang, and Aleix~M Martinez.
\newblock Recognition of action units in the wild with deep nets and a new
  global-local loss.
\newblock In {\em ICCV}, pages 3990--3999, 2017.

\bibitem{booth2018lsfm}
James Booth, Anastasios Roussos, Allan Ponniah, David Dunaway, and Stefanos
  Zafeiriou.
\newblock Large scale 3d morphable models.
\newblock {\em International Journal of Computer Vision}, 126(2-4):233--254,
  2018.

\bibitem{cao2018vggface2}
Qiong Cao, Li Shen, Weidi Xie, Omkar~M Parkhi, and Andrew Zisserman.
\newblock Vggface2: A dataset for recognising faces across pose and age.
\newblock In {\em 2018 13th IEEE International Conference on Automatic Face \&
  Gesture Recognition (FG 2018)}, pages 67--74. IEEE, 2018.

\bibitem{chen2019multi}
Zhao-Min Chen, Xiu-Shen Wei, Peng Wang, and Yanwen Guo.
\newblock Multi-label image recognition with graph convolutional networks.
\newblock In {\em Proceedings of the IEEE/CVF Conference on Computer Vision and
  Pattern Recognition}, pages 5177--5186, 2019.

\bibitem{LP8}
Ciprian Corneanu, Meysam Madadi, and Sergio Escalera.
\newblock Deep structure inference network for facial action unit recognition.
\newblock In {\em Proceedings of the European Conference on Computer Vision
  (ECCV)}, pages 298--313, 2018.

\bibitem{cui2020knowledge}
Zijun Cui, Tengfei Song, Yuru Wang, and Qiang Ji.
\newblock Knowledge augmented deep neural networks for joint facial expression
  and action unit recognition.
\newblock {\em Advances in Neural Information Processing Systems}, 33, 2020.

\bibitem{FACS}
Paul Ekman and Wallace Friesen.
\newblock Facial action coding system: A technique for the measurement of
  facial movement.
\newblock {\em Consulting Psychologists Press Palo Alto}, 12, 01 1978.

\bibitem{LSVM}
Rong-En Fan, Kai-Wei Chang, Cho-Jui Hsieh, Xiang-Rui Wang, and Chih-Jen Lin.
\newblock Liblinear: A library for large linear classification.
\newblock {\em J. Mach. Learn. Res.}, 9:1871--1874, 2008.

\bibitem{geng20193d}
Zhenglin Geng, Chen Cao, and Sergey Tulyakov.
\newblock 3d guided fine-grained face manipulation.
\newblock In {\em Proceedings of the IEEE/CVF Conference on Computer Vision and
  Pattern Recognition}, pages 9821--9830, 2019.

\bibitem{Resnet34}
Kaiming He, Xiangyu Zhang, Shaoqing Ren, and Jian Sun.
\newblock Deep residual learning for image recognition.
\newblock In {\em Proceedings of the IEEE conference on computer vision and
  pattern recognition}, pages 770--778, 2016.

\bibitem{jacob2021facial}
Geethu~Miriam Jacob and Bjorn Stenger.
\newblock Facial action unit detection with transformers.
\newblock In {\em Proceedings of the IEEE/CVF Conference on Computer Vision and
  Pattern Recognition}, pages 7680--7689, 2021.

\bibitem{jannat2020subject}
Sk~Rahatul Jannat, Diego Fabiano, Shaun Canavan, and Tempestt Neal.
\newblock Subject identification across large expression variations using 3d
  facial landmarks.
\newblock {\em arXiv preprint arXiv:2005.08339}, 2020.

\bibitem{F1}
L{\'a}szl{\'o}~A Jeni, Jeffrey~F Cohn, and Fernando De~La~Torre.
\newblock Facing imbalanced data--recommendations for the use of performance
  metrics.
\newblock In {\em 2013 Humaine association conference on affective computing
  and intelligent interaction}, pages 245--251. IEEE, 2013.

\bibitem{kelley1999iterative}
Carl~T Kelley.
\newblock {\em Iterative methods for optimization}.
\newblock SIAM, 1999.

\bibitem{kim2018deep}
Hyeongwoo Kim, Pablo Garrido, Ayush Tewari, Weipeng Xu, Justus Thies, Matthias
  Nie{\ss}ner, Patrick P{\'e}rez, Christian Richardt, Michael Zollh{\"o}fer,
  and Christian Theobalt.
\newblock Deep video portraits.
\newblock {\em TOG}, 37(4):1--14, 2018.

\bibitem{kollias2018aff}
Dimitrios Kollias and Stefanos Zafeiriou.
\newblock Aff-wild2: Extending the aff-wild database for affect recognition.
\newblock {\em arXiv preprint arXiv:1811.07770}, 2018.

\bibitem{kossaifi2019sewa}
Jean Kossaifi, Robert Walecki, Yannis Panagakis, Jie Shen, Maximilian Schmitt,
  Fabien Ringeval, Jing Han, Vedhas Pandit, Antoine Toisoul, Bjoern~W Schuller,
  et~al.
\newblock Sewa db: A rich database for audio-visual emotion and sentiment
  research in the wild.
\newblock {\em IEEE transactions on pattern analysis and machine intelligence},
  2019.

\bibitem{SRERL}
Guanbin Li, Xin Zhu, Yirui Zeng, Qing Wang, and Liang Lin.
\newblock Semantic relationships guided representation learning for facial
  action unit recognition.
\newblock In {\em Proceedings of the AAAI Conference on Artificial
  Intelligence}, volume~33, pages 8594--8601, 2019.

\bibitem{li2021write}
Lincheng Li, Suzhen Wang, Zhimeng Zhang, Yu Ding, Yixing Zheng, Xin Yu, and
  Changjie Fan.
\newblock Write-a-speaker: Text-based emotional and rhythmic talking-head
  generation.
\newblock In {\em Proceedings of the AAAI Conference on Artificial
  Intelligence}, volume~35, pages 1911--1920, 2021.

\bibitem{ROI}
W. Li, F. Abtahi, and Z. Zhu.
\newblock Action unit detection with region adaptation, multi-labeling learning
  and optimal temporal fusing.
\newblock In {\em 2017 IEEE Conference on Computer Vision and Pattern
  Recognition (CVPR)}, pages 6766--6775, Los Alamitos, CA, USA, July 2017. IEEE
  Computer Society.

\bibitem{li2018eac}
Wei Li, Farnaz Abtahi, Zhigang Zhu, and Lijun Yin.
\newblock Eac-net: Deep nets with enhancing and cropping for facial action unit
  detection.
\newblock {\em IEEE transactions on pattern analysis and machine intelligence},
  40(11):2583--2596, 2018.

\bibitem{MavadatiDISFA}
S.~Mohammad Mavadati, Mohammad~H. Mahoor, Kevin Bartlett, Philip Trinh, and
  Jeffrey~F. Cohn.
\newblock Disfa: A spontaneous facial action intensity database.
\newblock {\em IEEE Transactions on Affective Computing}, 4(2):151--160, 2013.

\bibitem{mollahosseini2017affectnet}
Ali Mollahosseini, Behzad Hasani, and Mohammad~H Mahoor.
\newblock Affectnet: A database for facial expression, valence, and arousal
  computing in the wild.
\newblock {\em IEEE Transactions on Affective Computing}, 10(1):18--31, 2017.

\bibitem{LP}
Xuesong Niu, Hu Han, Songfan Yang, Yan Huang, and Shiguang Shan.
\newblock Local relationship learning with person-specific shape regularization
  for facial action unit detection.
\newblock In {\em Proceedings of the IEEE Conference on Computer Vision and
  Pattern Recognition}, pages 11917--11926, 2019.

\bibitem{AU4EmoctionRecog}
Maja Pantic and Léon Rothkrantz.
\newblock Facial action recognition for facial expression analysis from static
  face images.
\newblock {\em Systems, Man, and Cybernetics, Part B: Cybernetics, IEEE
  Transactions on}, 34:1449 -- 1461, 07 2004.

\bibitem{paysan20093d}
Pascal Paysan, Reinhard Knothe, Brian Amberg, Sami Romdhani, and Thomas Vetter.
\newblock A 3d face model for pose and illumination invariant face recognition.
\newblock In {\em 2009 sixth IEEE international conference on advanced video
  and signal based surveillance}, pages 296--301. Ieee, 2009.

\bibitem{AUD4diagnose}
David~R Rubinow and Robert~M Post.
\newblock Impaired recognition of affect in facial expression in depressed
  patients.
\newblock {\em Biological psychiatry}, 31(9):947--953, 1992.

\bibitem{schroff2015facenet}
Florian Schroff, Dmitry Kalenichenko, and James Philbin.
\newblock Facenet: A unified embedding for face recognition and clustering.
\newblock In {\em Proceedings of the IEEE conference on computer vision and
  pattern recognition}, pages 815--823, 2015.

\bibitem{shao2021jaa}
Zhiwen Shao, Zhilei Liu, Jianfei Cai, and Lizhuang Ma.
\newblock Jaa-net: joint facial action unit detection and face alignment via
  adaptive attention.
\newblock {\em International Journal of Computer Vision}, 129(2):321--340,
  2021.

\bibitem{attnetionTrans2019}
Zhiwen Shao, Zhilei Liu, Jianfei Cai, Yunsheng Wu, and Lizhuang Ma.
\newblock Facial action unit detection using attention and relation learning.
\newblock {\em IEEE Transactions on Affective Computing}, 2019.

\bibitem{song2021uncertain}
Tengfei Song, Lisha Chen, Wenming Zheng, and Qiang Ji.
\newblock Uncertain graph neural networks for facial action unit detection.
\newblock In {\em Proceedings of the AAAI Conference on Artificial
  Intelligence}, volume~1, 2021.

\bibitem{song2021hybrid}
Tengfei Song, Zijun Cui, Wenming Zheng, and Qiang Ji.
\newblock Hybrid message passing with performance-driven structures for facial
  action unit detection.
\newblock In {\em Proceedings of the IEEE/CVF Conference on Computer Vision and
  Pattern Recognition}, pages 6267--6276, 2021.

\bibitem{vaswani2017attention}
Ashish Vaswani, Noam Shazeer, Niki Parmar, Jakob Uszkoreit, Llion Jones,
  Aidan~N Gomez, {\L}ukasz Kaiser, and Illia Polosukhin.
\newblock Attention is all you need.
\newblock In {\em Advances in neural information processing systems}, pages
  5998--6008, 2017.

\bibitem{vemulapalli2019compact}
Raviteja Vemulapalli and Aseem Agarwala.
\newblock A compact embedding for facial expression similarity.
\newblock In {\em Proceedings of the IEEE/CVF Conference on Computer Vision and
  Pattern Recognition}, pages 5683--5692, 2019.

\bibitem{wang2020region}
Kai Wang, Xiaojiang Peng, Jianfei Yang, Debin Meng, and Yu Qiao.
\newblock Region attention networks for pose and occlusion robust facial
  expression recognition.
\newblock {\em IEEE Transactions on Image Processing}, 29:4057--4069, 2020.

\bibitem{yang2021exploiting}
Huiyuan Yang, Lijun Yin, Yi Zhou, and Jiuxiang Gu.
\newblock Exploiting semantic embedding and visual feature for facial action
  unit detection.
\newblock In {\em Proceedings of the IEEE/CVF Conference on Computer Vision and
  Pattern Recognition}, pages 10482--10491, 2021.

\bibitem{yang2019facs3d}
Le Yang, Itir~Onal Ertugrul, Jeffrey~F Cohn, Zakia Hammal, Dongmei Jiang, and
  Hichem Sahli.
\newblock Facs3d-net: 3d convolution based spatiotemporal representation for
  action unit detection.
\newblock In {\em 2019 8th International Conference on Affective Computing and
  Intelligent Interaction (ACII)}, pages 538--544. IEEE, 2019.

\bibitem{yao2021one}
Guangming Yao, Yi Yuan, Tianjia Shao, Shuang Li, Shanqi Liu, Yong Liu, Mengmeng
  Wang, and Kun Zhou.
\newblock One-shot face reenactment using appearance adaptive normalization.
\newblock In {\em Proceedings of the AAAI Conference on Artificial
  Intelligence}, volume~35, pages 3172--3180, 2021.

\bibitem{you2020cross}
Renchun You, Zhiyao Guo, Lei Cui, Xiang Long, Yingze Bao, and Shilei Wen.
\newblock Cross-modality attention with semantic graph embedding for
  multi-label classification.
\newblock In {\em Proceedings of the AAAI Conference on Artificial
  Intelligence}, volume~34, pages 12709--12716, 2020.

\bibitem{zhang2021learning}
Wei Zhang, Xianpeng Ji, Keyu Chen, Yu Ding, and Changjie Fan.
\newblock Learning a facial expression embedding disentangled from identity.
\newblock In {\em Proceedings of the IEEE/CVF Conference on Computer Vision and
  Pattern Recognition}, pages 6759--6768, 2021.

\bibitem{zhang2021DLN}
Wei Zhang, Xianpeng Ji, Keyu Chen, Yu Ding, and Changjie Fan.
\newblock Learning a facial expression embedding disentangled from identity.
\newblock In {\em Proceedings of the IEEE/CVF Conference on Computer Vision and
  Pattern Recognition}, pages 6759--6768, 2021.

\bibitem{BP4D}
Xing Zhang, Lijun Yin, Jeffrey~F Cohn, Shaun Canavan, Michael Reale, Andy
  Horowitz, Peng Liu, and Jeffrey~M Girard.
\newblock Bp4d-spontaneous: a high-resolution spontaneous 3d dynamic facial
  expression database.
\newblock {\em Image and Vision Computing}, 32(10):692--706, 2014.

\bibitem{zhang2016multimodal}
Zheng Zhang, Jeff~M Girard, Yue Wu, Xing Zhang, Peng Liu, Umur Ciftci, Shaun
  Canavan, Michael Reale, Andy Horowitz, Huiyuan Yang, et~al.
\newblock Multimodal spontaneous emotion corpus for human behavior analysis.
\newblock In {\em Proceedings of the IEEE conference on computer vision and
  pattern recognition}, pages 3438--3446, 2016.

\bibitem{zhang2021flow}
Zhimeng Zhang, Lincheng Li, Yu Ding, and Changjie Fan.
\newblock Flow-guided one-shot talking face generation with a high-resolution
  audio-visual dataset.
\newblock In {\em Proceedings of the IEEE/CVF Conference on Computer Vision and
  Pattern Recognition}, pages 3661--3670, 2021.

\bibitem{AU4MicroExpression}
Guoying Zhao and Xiaobai Li.
\newblock Automatic micro-expression analysis: Open challenges.
\newblock {\em Frontiers in Psychology}, 10, 08 2019.

\bibitem{JPML}
K. {Zhao}, W.~S. {Chu}, F. {De la Torre}, J.~F. {Cohn}, and H. {Zhang}.
\newblock Joint patch and multi-label learning for facial action unit and
  holistic expression recognition.
\newblock {\em IEEE Transactions on Image Processing}, 25(8):3931--3946, 2016.

\bibitem{zhao2018learning}
Kaili Zhao, Wen-Sheng Chu, and Aleix~M Martinez.
\newblock Learning facial action units from web images with scalable weakly
  supervised clustering.
\newblock In {\em Proceedings of the IEEE Conference on computer vision and
  pattern recognition}, pages 2090--2099, 2018.

\bibitem{DRML}
Kaili Zhao, Wen-Sheng Chu, and Honggang Zhang.
\newblock Deep region and multi-label learning for facial action unit
  detection.
\newblock In {\em The IEEE Conference on Computer Vision and Pattern
  Recognition (CVPR)}, June 2016.

\bibitem{zhao2021robust}
Zengqun Zhao, Qingshan Liu, and Feng Zhou.
\newblock Robust lightweight facial expression recognition network with label
  distribution training.
\newblock In {\em Proceedings of the AAAI Conference on Artificial
  Intelligence}, volume~35, pages 3510--3519, 2021.

\bibitem{LP48}
Lin Zhong, Qingshan Liu, Peng Yang, Junzhou Huang, and {Dimitris N.} Metaxas.
\newblock Learning multiscale active facial patches for expression analysis.
\newblock {\em IEEE Transactions on Cybernetics}, 45(8):1499--1510, 8 2015.

\end{thebibliography}
}

\end{document}